\documentclass[conference]{IEEEtran}
\IEEEoverridecommandlockouts
\usepackage{cite}
\usepackage{amsmath,amssymb,amsfonts}
\usepackage{algorithmic}
\usepackage{graphicx}
\usepackage{textcomp}
\usepackage{xcolor}
\usepackage{booktabs} 
\usepackage{tabularx}
\usepackage{microtype}
\usepackage{amsmath}
\usepackage{amsfonts}
\usepackage{graphicx}
\usepackage{subfigure}
\usepackage{booktabs} 
\usepackage{tabularx}
\usepackage{arydshln}
\usepackage{algorithm}
\usepackage{algorithmic}

\def\BibTeX{{\rm B\kern-.05em{\sc i\kern-.025em b}\kern-.08em
    T\kern-.1667em\lower.7ex\hbox{E}\kern-.125emX}}
\begin{document}

\title{A Persona-based Multi-turn Conversation Model in an Adversarial Learning Framework} 


\author{\IEEEauthorblockN{Oluwatobi O. Olabiyi}
\IEEEauthorblockA{Capital One Conversation Research\\
Vienna, VA\\
Email: oluwatobi.olabiyi@capitalone.com}
\and
\IEEEauthorblockN{Anish Khazane}
\IEEEauthorblockA{Capital One Conversation Research\\
San Francisco, CA\\
Email: anish.khazane@capitalone.com}
\and
\IEEEauthorblockN{Erik T. Mueller}
\IEEEauthorblockA{Capital One Conversation Research\\
Vienna, VA\\
Email: erik.mueller@capitalone.com}}

\maketitle

\begin{abstract}
In this paper, we extend the persona-based sequence-to-sequence (Seq2Seq) neural network conversation model to
multi-turn dialogue by modifying the state-of-the-art hredGAN architecture. To achieve this, 
we introduce an additional input modality into the encoder and decoder of hredGAN to capture other 
attributes such as speaker identity, location, sub-topics, and other external attributes that 
might be available from the corpus of human-to-human interactions. The resulting persona hredGAN 
($phredGAN$) shows better performance than both the existing persona-based Seq2Seq and hredGAN models when 
those external attributes are available in a multi-turn dialogue corpus. 
This superiority is demonstrated on TV drama series with character consistency (such as Big Bang Theory 
and Friends) and customer service interaction datasets such as Ubuntu dialogue corpus in terms 
of perplexity, BLEU, ROUGE, and Distinct n-gram scores.
\end{abstract}

\section{Introduction}

Recent advances in machine learning especially with deep neural networks has led 
to tremendous progress in natural language processing and dialogue modeling research 
\cite{Sutskever2014, Vinyals2015, Serban2016}. Nevertheless, developing a good 
conversation model capable of fluent interactions between a human and a machine is still 
in its infancy. Most existing work relies on a limited dialogue 
history to produce responses with the assumption that the model parameters will 
capture all the modalities within a dataset. However, this is not true as dialogue corpora tend to be 
strongly multi-modal and practical neural network models find it difficult to disambiguate 
characteristics such as speaker personality, location, and sub-topic in the data.

Most work in this domain has primarily focused on optimizing dialogue consistency. For example, Serban et~al. \cite{Serban2016, Serban2017, Serban2017a} and Xing et~al. \cite{Xing} introduce a
Hierarchical Recurrent Encoder-Decoder (HRED) network architecture that combines a series of recurrent neural networks to capture long-term context state within a dialogue. However, the HRED system suffers from lack of diversity and does not support any guarantees on the generator output since the output conditional probability is not calibrated. Olabiyi et~al. \cite{Olabiyi2018} tackle this problem by training a modified HRED generator alongside an adversarial discriminator in order to provide a stronger guarantee to the generator's output. While the hredGAN system improves upon response quality, it does not capture speaker and other attributes modalities within a dataset and fails to generate persona-specific responses in datasets with multiple modalities.

At the same time, there has been some recent work on introducing persona into dialogue models. For example, Li et~al. \cite{Li2016b} integrate learnable speaker attributes into a single-turn (Seq2Seq) generative dialogue model. In this work, Li et~al. consider persona models: one with Speaker-only representation and another with Speaker and Addressee representations (Speaker-Addressee model), both of which capture aspects of speaker identity and interactions. Nguyen et~al. \cite{Nguyen2018} continue the same line of thought by considering a Seq2Seq dialogue model with Responder-only representations. In both cases, the attribute representation is learned during the system training. Zhang et~al. \cite{Zhang2018} propose a slightly different approach. Here, the attributes are a set of sentences describing the profile of the speaker. In this case, the attribute representation is not learned. The system however learns how to attend to different parts of the attributes during training. Still, the above persona-based models have a limited dialogue history (single-turn), suffer from exposure bias worsening the trade off between personalization and conversation quality, and cannot generate multiple responses given a dialogue context. This is evident in the relatively short and generic responses even though they generally capture the persona of the speaker.

\iftrue
\begin{table}[t]
\caption{Sample of $phredGAN$ outputs with discriminator score.}
\label{tb:dis_samples}
\begin{center}
\begin{tiny}
\setlength\tabcolsep{2.0pt} 
\begin{tabularx}{\linewidth}{ lllX }
\toprule
Source & Speaker & D(G(.)) & Utterance \\
\toprule
\multicolumn{4}{l}{\textbf{Big Bang Theory and Friends TV Series} } \\
Context\_0 & Penny & NA & You have to, we all have to be there at the same time to yell surprise!.\\
Context\_1 & Howard & NA & Okay, you have to understand something, we're in a hospital right now.\\
Response\_0 & Penny & 0.1705 & Okay, I'm sorry. I'm not really even like that.\\
Response\_1 & Penny & 0.1426 & Okay, I'm sorry, I'm sorry.\\
Response\_2 & Penny & 0.0688 & Okay. \\
\midrule
Context\_0 & Leonard & NA & Holy crap, Sheldon, did you just figure out a method for synthesizing a new stable super-heavy element?\\
Context\_1 & Sheldon & NA & Did I? Well, that can't be right. No one's ever done that before. Except me, because I just did it. Sheldon and his brain, yeah\\
Response\_0 & Leonard & 0.3976 & Sheldon , what are you doing ?\\
Response\_1 & Leonard & 0.3503 & Sheldon ?\\
\midrule
Context\_0 & Leonard & NA & She's gotten really hooked on Age of Conan, she's playing non-stop.\\
Response\_0 & Raj & 0.4890 & Okay , so we know , what about you and Penny doing here ?\\
Response\_1 & Raj & 0.3586 & Okay , so we would have to say that about people ?\\
Response\_2 & Raj & 0.1113 & Okay , let ' s go . \\
\toprule
\multicolumn{4}{l}{\textbf{UDC} } \\
Context\_0 & asker & NA & all i need some quick amarok help. i cant hear my music\\
Context\_1 & helper & NA & is amarok muted?\\
Context\_2 & asker & NA & no\\
Response\_0 & helper & 0.3079 & use the UNK drivers , and then run the " UNK " command to get the UNK \\
Response\_1 & helper & 0.1283 & what is the error message ? \\
Response\_2 & helper & 0.0725 & what version of ubuntu ? \\
\midrule
Context\_0 & asker & NA & anyone had problems with the kernel update from today? giving me a kernel panic\\
Context\_1 & helper & NA & you can select previous kernels at the bootloader (grub) menu on booth.\\
Response\_0 & asker & 0.3665 & it says that it is not installed . . . \\
Response\_1 & asker & 0.3195 & i'm not sure what i can find . . . \\
Response\_2 & asker & 0.0186 & it's a UNK , I'm not sure of the way . \\
\midrule
Context\_0 & asker & NA & how do I install Ubuntu?\\
Response\_0 & helper & 0.5797 & use the alternate cd , it should be in the repos , it's a good place to get the source of the kernel \\
Response\_1 & helper & 0.1984 & use the UNK package , it should work . . . \\
Response\_2 & helper & 0.0131 & use the UNK \\
\bottomrule

\end{tabularx}
\end{tiny}
\end{center}
\vskip -0.2in
\end{table}
\fi

To overcome these limitations, we propose $phredGAN$, a multi-modal $hredGAN$ dialogue system which additionally conditions the adversarial framework proposed by Olabiyi et~al. \cite{Olabiyi2018} on speaker and/or utterance attributes in order to maintain response quality of $hredGAN$ and still capture speaker and other modalities within a conversation. The attributes can be seen as another input modality as the utterance. The attribute representation is an embedding that is learned together with the rest of model parameters, similar to \cite{Li2016b}. The introduction of attributes allows the model to generate responses conditioned on particular attribute(s) across conversation turns. Since the attributes are discrete, it also allows for exploring what-if scenarios of model responses. We train and sample the proposed phredGAN similar to the procedure for $hredGAN$ \cite{Olabiyi2018}. To demonstrate model capability, we train on customer service related data such as the Ubuntu Dialogue Corpus (UDC) that is strongly bimodal between question poster and answerer, and character consistent TV scripts from two popular series, \textit{The Big Bang Theory} and \textit{Friends} with quantitative and qualitative analysis. We demonstrate system superiority over $hredGAN$ and the state-of-the-art persona conversational model in terms of dialogue response quality and quantitatively with perplexity, BLEU, ROUGE, and distinct n-gram scores.

The rest of the paper is organized as follows: In section \ref{model_arch}, we describe the model architecture. Section \ref{train_eval} describes model training and evaluation while section \ref{exp_res} contains experiments and result discussion. Finally, section \ref{concl} concludes and explores some future work directions. 

\section{Model Architecture}
\label{model_arch}
In this section, we briefly introduce the state-of-the-art $hredGAN$ model and subsequently show how we derive the persona version by combining it with the distributed representation of the dialogue speaker and utterance attributes.

\subsection{$hredGAN$: Adversarial Learning Framework}

The $hredGAN $proposed by Olabiyi et.~al \cite{Olabiyi2018} contains three major components. 

\textbf{Encoder}: The encoder consists of three RNNs, the $aRNN$ that encodes an utterance for attention memory, the $eRNN$ that encodes an utterance for dialogue context, and the $cRNN$ that encodes a multi-turn dialogue context from the $eRNN$ outputs. The final states of $aRNN$ and $cRRN$ are concatenated to form the final encoder state. 

\textbf{Generator}: The generator contains a single decoder RNN, $dRNN$, initialized with the final encoder state. The $dRNN$ inputs consists of $cRNN$ output, embedding of the ground truth or previously generated tokens, as well as noise samples and the attention over the $aRNN$ outputs.

\textbf{Discriminator}: The discriminator also contains a single RNN, $D_{RNN}$, initialized by the final encoder state. In the case of $hredGAN$ \cite{Olabiyi2018}, it is a bidirectional RNN that discriminates at the word level to capture both the syntactic and semantic difference between the ground truth and the generator output.

\textbf{Problem Formulation}:         
The $hredGAN$  \cite{Olabiyi2018} formulates multi-turn dialogue response generation as: given a dialogue history of sequence of utterances, $\boldsymbol{X_i}=\big(X_1,X_2,\cdots,X_i\big)$, where each utterance $X_i=\big(X_i^1,X_i^2,\cdots,X_i^{M_i }\big)$ contains a variable-length sequence of $M_i$ word tokens such that ${X_i}^j \in V$ for vocabulary $V$, the dialogue model produces an output $Y_i=\big(Y_i^1,Y_i^2,\cdots,Y_i^{T_i}\big)$, where $T_i$ is the
number of generated tokens. The framework uses a conditional GAN structure to learn a mapping from an observed dialogue history to a sequence of output tokens. The generator, $G$, is trained to produce sequences that cannot be distinguished from the ground truth by an adversarially trained discriminator, $D$, akin to a two-player min-max optimization problem. The generator is also trained to minimize the cross-entropy loss $\mathcal{L}_{MLE}(G)$ between the ground truth $X_{i+1}$, and the generator output $Y_i$. The following objective summarizes both goals:

\begin{multline} \label{eq:mlegan}
G^*, D^* = arg\mathop{min}\limits_G\mathop{max}\limits_D\big(\lambda_{G}\mathcal{L}_{cGAN}(G,D) + \\
\lambda_{M}\mathcal{L}_{MLE}(G)\big).
\end{multline}

where $\lambda_{G}$ and $\lambda_{M}$ are hyperparameters and $\mathcal{L}_{cGAN}(G,D)$ and $\mathcal{L}_{MLE}(G)$ are defined in Eqs. (5) and (7) of \cite{Olabiyi2018} respectively.

Note that the generator $G$ and discriminator $D$ share the same encoder and embedding representation of the word tokens.

\begin{figure*}[ht]
\begin{center}
\centerline{\includegraphics[width=0.85\textwidth]{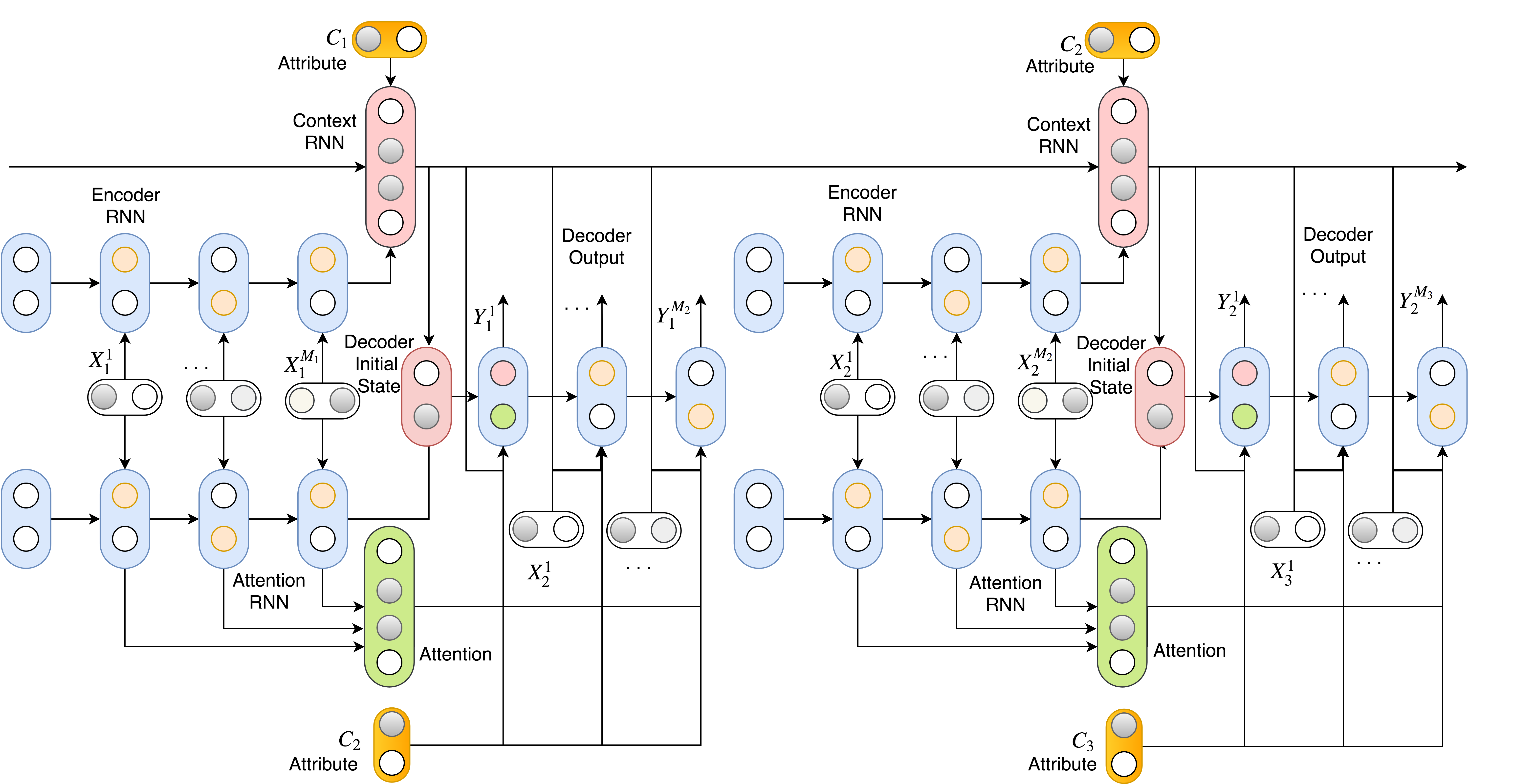}}
\caption{\textbf{The PHRED generator with local attention -}
\textbf{Encoder}: Encoder RNN, $eRNN$, Attention RNN, $aRNN$ and the Context RNN, $cRRN$. \textbf{Generator}: Decoder RNN, $dRNN$. The same encoder is shared between the generator and the discriminator (Fig. 1 of \cite{Olabiyi2018}). The attributes C allow the generator to condition its response on the utterance attributes such as
speaker identity, subtopics, and so on.}
\label{chred_att}
\end{center}
\vskip -0.2in
\end{figure*}

\subsection{$phredGAN$: Persona Adversarial Learning Framework}

The proposed architecture of $phredGAN$ is very similar to that of $hredGAN$ summarized above. The only difference is that the dialogue history is now 
$\boldsymbol{X_i}=\big((X_1,C_1),(X_2,C_2),\cdots,(X_i,C_i)\big)$ where $C_i$ is additional input that represents the speaker and/or utterance attributes. Note that 
$C_i$ can either be a sequence of tokens or a single token such that ${C_i}^j \in Vc$ for vocabulary $Vc$. The embedding for attribute tokens is also learned similar 
to that of word tokens. The modified system is as follows: 

\textbf{Encoder}: In addition to the three RNNs in the encoder of $hredGAN$, if the attribute $C_i$ is a sequence of tokens, then another RNN, $sattRNN$ is used to summarize the token embeddings; otherwise a single attribute embedding is concatenated with the output of $eRNN$ as shown in Fig. \ref{chred_att}. 

\textbf{Generator}: In addition to the $dRNN$ in the generator of $hredGAN$, if the attribute $C_{i+1}$ is a sequence of tokens, then another RNN, $tattRNN$ is used to summarize the attribute token embeddings; otherwise a single attribute embedding is concatenated with the other inputs of $dRNN$ as shown in Fig. \ref{chred_att}.

\textbf{Discriminator}: In addition to the $D_{RNN}$ in the discriminator of $hredGAN$, if the attribute $C_{i+1}$ is a sequence of tokens, then the same $tattRNN$ is used to summarize the attribute token embeddings; otherwise the single attribute embedding is concatenated with the other inputs of $D_{RNN}$ in Fig. 1 of \cite{Olabiyi2018}.

\textbf{Noise Injection}: Although \cite{Olabiyi2018} demonstrated that injecting noise at the word level seems to perform better than at the utterance level for $hredGAN$, we found that this is dataset-dependent for $phredGAN$. The $phredGAN$ model with utterance-level noise injection and word-level noise injection are tagged $phredGAN_u$ and $phredGAN_w$ respectively.  

The losses, $\mathcal{L}_{cGAN}(G,D)$ and $\mathcal{L}_{MLE}(G)$ in eq. (\ref{eq:mlegan}) are then respectively updated as:

\begin{multline} \label{eq:cgan}
\mathcal{L}_{cGAN}(G,D) = \mathbb{E}_{\boldsymbol{X_i},C_{i+1},X_{i+1}}[\log{}D(\boldsymbol{X_i},C_{i+1},X_{i+1})] + \\
\mathbb{E}_{\boldsymbol{X_i},C_{i+1},Z_i}[1-\log{}D(\boldsymbol{X_i},C_{i+1},G(\boldsymbol{X_i},C_{i+1},Z_i))]
\end{multline}

\begin{equation} \label{eq:mle}
\mathcal{L}_{MLE}(G) = \mathbb{E}_{X_{i+1}}[-log~P_G\big(X_{i+1}|\boldsymbol{X_i},C_{i+1},Z_i\big)].
\end{equation}

The addition of speaker or utterance attributes allows the dialogue model to exhibit personality traits giving consistent responses across style, gender, location, and so on.

\iftrue
\begin{algorithm}[h]
  \scriptsize
\begin{algorithmic}
\caption{Adversarial Learning of phredGAN}
   \label{alg:chredgan}
   \REQUIRE A generator $G$ with parameters $\theta_G$. 
   \REQUIRE A discriminator $D$ with parameters $\theta_{D}$.
   \REQUIRE Training hyperparameters, $\lambda_{G}$, and $\lambda_{M}$.
   \FOR {number of training iterations}
   \STATE Initialize $cRNN$ to zero\_state, $\boldsymbol{h_0}$
   \STATE Sample a mini-batch of conversations, $\boldsymbol{X} = \{X_i, C_i\}_{i=1}^N$, 
   $\boldsymbol{X_i}=\big((X_1,C_1),(X_2,C_2),\cdots,(X_i,C_i)\big)$ with $N$ utterances. 
   Each utterance mini batch $i$ contains $M_i$ word tokens.
   \FOR{$i=1$ {\bfseries to} $N-1$}
   \STATE Update the context state.
   \STATE $\boldsymbol{h_i} = cRNN(eRNN(E(X_i)),\boldsymbol{h_{i-1}}, C_i)$
   \STATE Compute the generator output similar to Eq. (11) in \cite{Olabiyi2018}.
   \begin{align*}\abovedisplayskip=0pt \belowdisplayskip=0pt
   P_{\theta_G}\big(Y_i|,Z_i,\boldsymbol{X_i}, C_{i+1}\big) &= \big\{P_{\theta_G}\big(Y_i^j|X_{i+1}^{1:j-1},Z_i^j,\boldsymbol{X_i}, C_{i+1}\big)\big\}_{j=1}^{M_{i+1}}
   \end{align*}
   \STATE Sample a corresponding mini batch of utterance $Y_i$.
   \STATE $Y_i \sim P_{\theta_G}\big(Y_i|,Z_i,\boldsymbol{X_i}, C_{i+1}\big)$ 
   \ENDFOR
   \STATE Compute the adversarial discriminator accuracy $D_{acc}$ over $N-1$ utterances
   $\{Y_i\}_{i=1}^{N-1}$ and $\{X_{i+1}\}_{i=1}^{N-1}$
   \IF{$D_{acc} < acc_{D_{th}}$}
   \STATE Update $phredGAN$'s $\theta_{D}$ with gradient of the discriminator loss.
   \STATE $\sum\limits_{i}[\nabla_{\theta_{D}}\log{}D(\boldsymbol{h_i}, C_{i+1}, X_{i+1}) + \nabla_{\theta_{D}}\log\big(1-D(\boldsymbol{h_i}, C_{i+1}, Y_i)\big)]$
   \ENDIF
   \IF{${D}_{acc} < acc_{G_{th}}$}
   \STATE Update $\theta_G$ with the generator's MLE loss only.
   \STATE $\sum\limits_{i}[\nabla_{\theta_G}-\log{}P_{\theta_G}\big(Y_i|,Z_i,\boldsymbol{X_i}, C_{i+1}\big)]$
   \ELSE
   \STATE Update $\theta_G$ with attribute,  adversarial and MLE losses.
   \STATE $\sum\limits_{i}[\lambda_{G}\nabla_{\theta_G}\log{}D(\boldsymbol{h_i}, C_{i+1}, Y_i) 
                  + \lambda_{M}\nabla_{\theta_G}{-\log{}}P_{\theta_G}\big(Y_i|,Z_i,\boldsymbol{X_i}, C_{i+1}\big)]$
   \ENDIF
   \ENDFOR
\end{algorithmic}
\end{algorithm}
\fi

\section{Model Training and Inference}
\label{train_eval}

\subsection{Model Training}
We train both the generator and the discriminator (with a shared encoder) of $phredGAN$ using the same training procedure in Algorithm 1 
with $\lambda_G = \lambda_M = 1$ \cite{Olabiyi2018}. Since the encoder, word embeddings and attribute embeddings are shared, we are able to train the system end-to-end with back-propagation. 

\textbf{Encoder}: The encoder RNNs, $eRNN$, and $eRNN$ are bidirectional while $cRRN$ is unidirectional. All RNN units are 3-layer GRU cells with a hidden state size of 512.
We use a word vocabulary size, $V = 50,000$, with a word embedding size of 512. The number of attributes, $Vc$ is dataset-dependent but we use 
an attribute embedding size of 512. In this study, we only use one attribute per utterance, so there is no need to use RNNs, $sattRNN$ and $tattRNN$ to combine the attribute embeddings.

\textbf{Generator}: The generator RNN, $dRNN$ is also a 3-layer GRU cell with a hidden state size of 512. The $aRNN$ outputs are connected to the $dRNN$ input using an additive attention mechanism \cite{Bahdanau2015}.

\textbf{Discriminator}: The discriminator RNN, $D_{RNN}$, is a bidirectional RNN, each 3-layer GRU cell having a hidden state size of 512. 
The output of both the forward and the backward cells for each word are 
concatenated and passed to a fully-connected layer with binary output. The output is 
the probability that the word is from the ground truth given the past and future words of the sequence as well as the responding speaker's embedding.

\textbf{Others}:
All parameters are initialized with Xavier uniform random initialization \cite{Glorot2010}. 
Due to the large word vocabulary size, we use sampled softmax loss \cite{Jean2015} for MLE loss to expedite the training process. However, we use full softmax
for model evaluation.
The parameter update is conditioned on the discriminator accuracy performance as in \cite{Olabiyi2018} with $acc_{D_{th}} = 0.99$ and $acc_{G_{th}} = 0.75$. 
The model is trained end-to-end using the stochastic gradient descent algorithm.  
Finally, the model is implemented, trained, and evaluated using the TensorFlow deep learning framework.

\subsection{Model Inference}    
We use an inference strategy similar to the approach in Olabiyi et. al \cite{Olabiyi2018}. The only differences between training and inference are :
(i) The generator is run in autoregressive mode with greedy decoding by passing the previously generated word token to the input of the $dRNN$ at the next step.
(ii) A modified noise sample $\mathcal{N}(0,\alpha\boldsymbol{I})$ is passed into the generator input.

For the modified noise sample, 
we perform a linear search for $\alpha$ with sample size $L=1$ based on the average discriminator loss, $-logD(G(.))$ \cite{Olabiyi2018} using trained models run in 
autoregressive mode to reflect performance in actual deployment. The optimum $\alpha$ value is then used for all inferences and evaluations.
During inference, we condition the dialogue response generation on the encoder outputs, noise samples, word embedding, and the attribute embedding of the intended responder. With multiple noise samples, $L=64$, we rank the generator outputs by the discriminator which is also conditioned on the encoder outputs, and the intended responder's embedding. The final response is the response ranked highest by the discriminator.

\section{Experiments and Results}
\label{exp_res}
In this section, we explore $phredGAN$'s results on two conversational datasets and compare its performance to the persona system in Li et al. \cite{Li2016b} 
and $hredGAN$ \cite{Olabiyi2018} in terms of quantitative and qualitative measures.

\subsection{Datasets}
\textbf{TV Series Transcripts} dataset \cite{Serban2016}. We train our model on transcripts from the two popular TV drama series, Big Bang Theory and Friends. Following a preprocessing setup similar to \cite{Li2016b}, we collect utterances from the top 12 speakers from both series to construct a corpus of 5,008 lines of multi-turn dialogue. We split the corpus into training, development, and test sets with 94\%, 3\%, and 3\% proportions, respectively, and pair each set with a corresponding attribute file that maps speaker IDs to utterances in the combined dataset.

Due to the small size of the combined transcripts dataset, we first train our model on the larger Movie Triplets Corpus (MTC) by Banchs et al. \cite{Banchs2012} which consists of 240,000 dialogue triples. We pre-train our model on this dataset to initialize our model parameters to avoid overfitting on a relatively small persona TV series dataset. After pre-training on MTC, we reinitialize the attribute embeddings in the generator from a uniform distribution following a Xavier initialization \cite{Glorot2010} for training on the combined person TV series dataset.

\textbf{Ubuntu Dialogue Corpus} (UDC) dataset \cite{Serban2017}. We train our model on 1.85 million conversations of multi-turn dialogue from the Ubuntu community hub, with an average of 5 utterances per conversation. We assign two types of speaker IDs to utterances in this dataset: questioner and helper. We follow the same training, development, and test split as the UDC dataset in \cite{Olabiyi2018}, with 90\%, 5\%, and 5\% proportions, respectively.

While the overwhelming majority of utterances in UDC follow two speaker types, the dataset does include utterances that are not classified under either a questioner or helper speaker type. To remain consistent, we assume that there are only two speaker types within this dataset and that the first utterance of every dialogue is from a questioner. This simplifying assumption does introduce a degree of noise into the model's ability to construct attribute embeddings. However, our experimental results demonstrate that our model is still able to differentiate between the larger two speaker types in the dataset. 

\begin{table}[t]
\caption{$predGAN$ vs. Seq2Seq Persona Models \cite{Li2016b} Performance.}
\label{tb:perp}
\begin{center}
\begin{small}
\begin{sc}
\begin{tabular}{llll}
\toprule
Metric & SM\footnotemark\cite{Li2016b} & SAM\cite{Li2016b} & $phredGAN$   \\
\midrule
\textbf{TV Series}\\
Perplexity  & \textbf{25.0} & 25.4 & 25.9 \\
BLEU-4     & 1.88\%       & 1.90\% & \textbf{3.00\%} \\
ROGUE-2    & - & - & \textbf{0.4044} \\
Distinct-1 & - & - & \textbf{0.1765} \\
Distinct-2 & - & - & \textbf{0.2164} \\
\bottomrule
\end{tabular}
\end{sc}
\end{small}
\end{center}
\vspace{-20pt}
\end{table}

\footnotetext{SM stands for Speaker Model and SAM stands for Speaker-Addressee Model}

\subsection{Evaluation Metrics}

We use similar evaluation metrics as in \cite{Olabiyi2018} including perplexity, BLEU \cite{Papineni2002}, ROUGE \cite{Lin2014}, 
and distinct n-gram \cite{Li2016} scores.

\subsection{Baseline}

We compare our system to \cite{Li2016b} which uses a Seq2Seq framework in conjunction with learnable persona embeddings. Their work explores two persona models to incorporate vector representations of speaker interaction and speaker attributes into the decoder of their Seq2Seq model, i.e., Speaker and Speaker-Addressee models. While we compare with both models quantitatively, we mostly compare with the Speaker-Addressee model qualitatively. Our quantitative comparison uses perplexity and BLEU-4 scores as those are the ones reported in \cite{Li2016b}. In addition, we also measure our model performance in terms of ROGUE and Distinct n-gram scores for the purpose of completeness. For fair comparison, we use the same TV drama series dataset used in their study.

We also compare our system to $hredGAN$ from \cite{Olabiyi2018} in terms of perplexity, ROGUE, and distinct n-grams scores. In \cite{Olabiyi2018}, the authors recommend the version with word-level noise injection, $hredGAN_w$, so we use this version in our comparison. Also for fair comparison, we use the same UDC dataset as reported in \cite{Olabiyi2018}. The only addition we made is to add the speaker attribute to the utterances of the dataset as described in the Dataset subsection.

\subsection{Hyperparameter search}

Prior to evaluation, we determine the noise injection method and the optimum noise variance $\alpha$ that are suitable for the two datasets. We consider the two variations of $phredGAN$, that is, $phredGAN_u$ for utterance-level and $phredGAN_w$ for word-level noise injections. We notice a partial mode collapse with $phredGAN_w$ on the combined TV transcripts, likely due to the high variation of word-level perturbation on a very limited dataset. However, this issue was rectified by $phredGAN_u$. Therefore, we use $phredGAN_u$ for the combined TV series dataset and $phredGAN_w$ for the UDC dataset. We perform a linear search for optimal noise variance values between 1 and 30 at an increment of 1, with a sample size of $L=1$. We obtained an optimal $\alpha$ of 2 with $phredGAN_u$ for the combined TV series dataset and an optimal $\alpha$ of 5 for $phredGAN_w$ for the much larger UDC dataset.

\begin{table}[t]
\caption{$phredGAN$ vs. $hredGAN$ \cite{Olabiyi2018} Performance}
\label{tb:perp1}
\begin{center}
\begin{small}
\begin{sc}
\begin{tabular}{lll}
\toprule
Metric & $hredGAN$ \cite{Olabiyi2018}& $phredGAN$ \\
\midrule
\textbf{UDC}\\
Perplexity      & 48.18      & \textbf{27.3}     \\
ROUGE-2 (F1)        & 0.1252 & \textbf{0.1692} \\
DISTINCT-1 & 14.05\% & \textbf{20.12\%} \\
DISTINCT-2 & \textbf{31.24\%} & 24.53\%\\

\bottomrule
\end{tabular}
\end{sc}
\end{small}
\end{center}
\vspace{-20pt}
\end{table}

\subsection{Results}
After training phredGAN models on the TV series and UDC datasets, we ran inference on some example dialogue contexts. The responses and their discriminator scores from $phredGAN$ are listed in Table \ref{tb:dis_samples}. The table shows that $phredGAN$ (i) can handle multi-turn dialogue context with utterances and corresponding persona attributes; (ii) generates responses conditioned on a persona attribute; (iii) generates multiple responses per dialogue context and scores their human likelihood by the discriminator. We observe that the discriminator score is generally reasonable with longer, more informative, and more persona-related responses receiving higher scores. It is worth noting that this behavior, although similar to the behavior of a human judge, is learned without supervision. Furthermore, we observe that $phredGAN$ responses retain contextual consistency, sometimes referencing background information inherent in the conversation between two speakers. For example, in the second sample of the TV series in Table \ref{tb:dis_samples}, $phredGAN$ generator, Leonard refers to Sheldon by name. Also, in the third sample, $phredGAN$, Raj refers to Penny when responding to Leonard who happens to be Penny's boy friend. We observe similar persona-based response generation for the UDC dataset with distinct communication styles between the asker and the helper.

We will now present performance comparisons of $phredGAN$ against the baselines, $hredGAN$, and Li et al.'s persona Seq2Seq models. 

\subsubsection{Quantitative Analysis}

We first report the performance of $phredGAN_u$ on TV series transcripts in table \ref{tb:perp}. Our system actually performs slightly worse than both variations of the Speaker Model and Speaker-Addressee systems in \cite{Li2016b} in terms of the perplexity measure. This is because the entropy of multi-turn dialogue is higher than that of single-turn. Similar observations have been made by Serban et al. \cite{Serban2016} about seq2seq and HRED dialogue models.  However, $phredGAN_u$ gives a significantly larger BLEU-4 score than both Speaker and Speaker-Addressee models. We attribute this improvement to (i) the multi-turn dialogue context, and (ii) training in an adversarial framework, which forces the model to produce longer, more informative, and diverse responses that have high topic relevance even with a limited dataset. Also, unlike Speaker-Addressee models which suffer from lower response quality due to persona conditioning, conditioning the generator and discriminator of $phredGAN$ on speaker embeddings does not compromise the system's ability to produce diverse responses. This problem might have been alleviated by the adversarial training too.

We also compare $phredGAN$ with the recommended variant of $hredGAN$ that includes word-level noise injection at the decoder on the UDC dataset. The results are summarized in Table \ref{tb:perp1}. We note an improvement in a variety of evaluation metrics including perplexity, ROUGE, and distinct n-grams, with the exception of distinct 2-grams. This is expected as $phredGAN$ should be generally less diverse than $hredGAN$ since the number of distinct data distribution modes is more for the $phredGAN$ dataset due to the persona attributes. However, this leads to better response quality with persona, something not achievable with $hredGAN$. Also, the better ROUGE(f1) score indicates that $phredGAN$ is able to strike a better balance between diversity and precision while still capturing the characteristics of the speaker attribute modality in the UDC dataset.

\subsubsection{Qualitative Analysis}

A qualitative assessment of these results are in Table \ref{tb:samples} with responses from several characters in the TV series dataset and the two characters in UDC. 

We see that for TV drama series, $phredGAN$ responses are comparatively more informative than that of the Speaker-Addressee model of \cite{Li2016b}. For example, with Speaker-Addressee model, nearly all the characters in the TV series respond with ``Of course I love you.'' to the dialogue context, ``Do you love me?" despite the fact that some of the responders sometimes have unfriendly relationship with the addressee. Many of the novel situations explored by $phredGAN$ are unachievable with the Speaker-Addressee model due to a lack of informative responses. For example, by conditioning as Sheldon from The Big Bang Theory and asking ``Do you like me?", our model responds with annoyance if conditioned as Penny (``No, you don't understand. You're an idiot"), brevity with Leonard (``Yes?"), and sarcasm with Raj (``Well , you know , we could be a little more than my friend's friends.") The wide range of responses indicate our model's ability to construct distinct attribute embeddings for each character even from a limited dataset. The other interesting responses in Table \ref{tb:samples} indicate our model's ability to infer not only the context of the conversation but important character information about the Addressee.

We also see similar results with our model's output on UDC. We demonstrate that by conditioning as either a helper or questioner from the UDC dataset, our model is able to respond differently to input utterances as well as stay close to the context of the conversation.

\iftrue
\begin{table}[t]
\caption{ phredGAN sample responses on TV Series and UDC.}
\label{tb:samples}
\begin{center}
\begin{tiny}
\setlength\tabcolsep{1.0pt} 
\begin{tabularx}{\linewidth}{ ll X }
\toprule
Model & Speaker & Response \\
\toprule
\multicolumn{2}{l}{\textbf{Big Bang Theory Series} } \\
\textbf{Context} & Sheldon & Do you like me?\\
\textbf{phredGAN} & Leonard & Yes ?\\
\textbf{phredGAN} & Penny & No , you don ' t understand . You ' re an idiot . \\
\textbf{phredGAN} & Howard & Well , I don ' t know what to say . This is the part I ' ve ever seen .\\
\textbf{phredGAN} & Raj &  Well , you know , we could be a little more than my friends ' s friends .\\
\hline
\textbf{Context} & Emily & Do you like me?\\
\textbf{Speaker-Addressee} & Sheldon & Of course, I love you, Emily \\
\textbf{Speaker-Addressee} & Penny & Of course, I love you, Emily.\\
\textbf{Speaker-Addressee} & Rachel & Of course I love you.\\
\midrule
\textbf{Context\_0} & Howard & We can’t go home. \\
\textbf{Context\_1} & Leonard& Alright, I'll be right back. \\
\textbf{phredGAN} & Howard& Okay , well , before you and Penny get out and get married with Penny just get married. \\
\midrule 
\textbf{Context\_0} & Amy & Spent a lot of my childhood throwing coins into wishing wells hoping for friends \\
\textbf{Context\_1} & Penny & Whoa, whoa, give me a second to catch up here. Hold on \\
\textbf{Context\_2} & Amy & That'd be my boyfriend. Happier playing his dopey Star Trek game with his friends than hanging out with me. \\
\textbf{phredGAN} & Penny & You know , when you ' re being so nice about Sheldon and I want to take him to like his own girlfriend \\
\midrule
\textbf{Context} & Sheldon & I recently read that during World War Two, Joseph Stalin had a research program to create supersoldiers by having women impregnated by gorillas. \\
\textbf{predGAN} & Howard & You ' re right . I don ' t know what you ' re talking about , Sheldon . \\
\toprule
\multicolumn{2}{l}{\textbf{Friends TV Series} } \\
\textbf{Context} & Rachel & Do you like me?\\
\textbf{phredGAN} & Joey & Yeah .\\
\textbf{phredGAN} & Ross & Well , no , no , no . I mean , I just , I don ' t know . . . \\
\textbf{phredGAN} & Monica & Well , I do . And I ' m not gonna tell you another guy .\\
\textbf{phredGAN} & Chandler & No , I do not . You are a mean little man . .\\
\hline
\textbf{Context} & Rachel & Do you like me?\\
\textbf{Speaker-Addressee} & Chandler & Of course I love you. \\
\toprule
\textbf{Ubuntu} \\
\textbf{Context} & Asker & i need to install ubuntu on a pendrive so i can install software on the pendrive after booting from the usb, is it possible?\\
\textbf{phredGAN} & Helper & you can just boot into the live cd and install grub. \\
\midrule
\textbf{Context} & Asker & anyone had problems with the kernel update from today? giving me a kernel panic there a simple way to make a rollback? anyone if I can go back to my previous kernel?\\
\textbf{phredGAN} & Helper & what version of ubuntu are you using ? \\
\midrule
\textbf{Context} & Helper & did you see my message about looking at other log files?\\
\textbf{predGAN} & Asker & yeah , it ' s not that I ' m trying to get the UNK to work \\
\midrule
\textbf{Context} & Asker & you can select previous kernels at the bootloader (grub) menu on boot\\
\textbf{phredGAN} & Helper & i ' m not to be able to get the grub to the grub menu \\
\bottomrule
\end{tabularx}
\end{tiny}
\end{center}
\vspace{-20pt}
\end{table}
\fi

\section{Conclusion and Future Work}
\label{concl}

In this article, we improve upon state-of-the-art persona-based response generation models by adding attribute embeddings to $hredGAN$ in order to capture a representation of speaker identity and style in a multi-turn dialogue. Our model outperforms the existing Speaker-Model and Speaker-Addressee models from \cite{Li2016b} with respect to BLEU score and even improves upon $hredGAN$ with respect to ROUGE, distinct n-grams, and perplexity scoring. We also see a qualitative improvement in response quality that more clearly separates speaker identity from even limited datasets.

In the future, we hope to carry out a human evaluation to confirm qualitative improvement of our model's outputs. We also hope to extend this work to Multi-class GAN by optimizing over two separate adversarial losses, one for speaker attribute classification and the other for dialogue generation real/fake discrimination. This will allow us to further improve on persona distinctions without the loss of response quality.

\bibliographystyle{IEEEtran}
\bibliography{IEEEabrv,conference_041818}

\begin{thebibliography}{10}
\providecommand{\url}[1]{#1}
\csname url@samestyle\endcsname
\providecommand{\newblock}{\relax}
\providecommand{\bibinfo}[2]{#2}
\providecommand{\BIBentrySTDinterwordspacing}{\spaceskip=0pt\relax}
\providecommand{\BIBentryALTinterwordstretchfactor}{4}
\providecommand{\BIBentryALTinterwordspacing}{\spaceskip=\fontdimen2\font plus
\BIBentryALTinterwordstretchfactor\fontdimen3\font minus
  \fontdimen4\font\relax}
\providecommand{\BIBforeignlanguage}[2]{{%
\expandafter\ifx\csname l@#1\endcsname\relax
\typeout{** WARNING: IEEEtran.bst: No hyphenation pattern has been}%
\typeout{** loaded for the language `#1'. Using the pattern for}%
\typeout{** the default language instead.}%
\else
\language=\csname l@#1\endcsname
\fi
#2}}
\providecommand{\BIBdecl}{\relax}
\BIBdecl

\bibitem{Sutskever2014}
I.~Sutskever, O.~Vinyals, and Q.~Le, ``Sequence to sequence learning with
  neural networks,'' in \emph{Proceedings of Advances in Neural Information
  Processing Systems (NIPS)}, 2014, pp. 3104–--3112.

\bibitem{Vinyals2015}
O.~Vinyals and Q.~Le, ``A neural conversational model,'' in \emph{Proceedings
  of ICML Deep Learning Workshop}, 2015.

\bibitem{Serban2016}
I.~Serban, A.~Sordoni, Y.~Bengio, A.~Courville, and J.~Pineau, ``Building
  end-to-end dialogue systems using generative hierarchical neural network
  models,'' in \emph{Proceedings of The Thirtieth AAAI Conference on Artificial
  Intelligence (AAAI 2016)}, 2016, pp. 3776--3784.

\bibitem{Serban2017}
I.~V. Serban, A.~Sordoni, R.~Lowe, L.~Charlin, J.~Pineau, A.~Courville, and
  Y.~Bengio, ``A hierarchical latent variable encoder-decoder model for
  generating dialogue,'' in \emph{Proceedings of The Thirty-first AAAI
  Conference on Artificial Intelligence (AAAI 2017)}, 2017.

\bibitem{Serban2017a}
I.~V. Serban, T.~Klinger, G.~Tesauro, K.~Talamadupula, B.~Zhou, Y.~Bengio, and
  A.~Courville, ``Multiresolution recurrent neural networks: An application to
  dialogue response generation,'' in \emph{Proceedings of The Thirty-first AAAI
  Conference on Artificial Intelligence (AAAI 2017)}, 2017.

\bibitem{Xing}
C.~Xing, W.~Wu, Y.~Wu, M.~Zhou, Y.~Huang, and W.~Ma, ``Hierarchical recurrent
  attention network for response generation,'' in \emph{arXiv preprint
  arXiv:1701.07149}, 2017.

\bibitem{Olabiyi2018}
O.~Olabiyi, A.~Salimov, A.~Khazane, and E.~Mueller, ``Multi-turn dialogue
  response generation in an adversarial learning framework,'' in \emph{arXiv
  preprint arXiv:1805.11752}, 2018.

\bibitem{Li2016b}
J.~Li, M.~Galley, C.~Brockett, G.~Spithourakis, J.~Gao, and B.~Dolan, ``A
  persona-based neural conversation model,'' in \emph{Proceedings of the 54th
  Annual Meeting of the Association for Computational Linguistics}, 2016, pp.
  994–--1003.

\bibitem{Nguyen2018}
H.~Nguyen, D.~Morales, and T.~Chin, ``A neural chatbot with personality,'' in
  \emph{Stanford NLP Course website:
  https://web.stanford.edu/class/cs224n/reports/2761115.pdf}, 2018.

\bibitem{Zhang2018}
S.~Zhang, E.~Dinan, J.~Urbanek, A.~Szlam, D.~Kiela, and J.~Weston,
  ``Personalizing dialogue agents: I have a dog, do you have pets too?'' in
  \emph{arXiv preprint arXiv:1801.07243v3}, 2018.

\bibitem{Bahdanau2015}
D.~Bahdanau, K.~Cho, and Y.~Bengio, ``Neural machine translation by jointly
  learning to align and translate,'' in \emph{Proceedings of International
  Conference of Learning Representation (ICLR 2015)}, 2015.

\bibitem{Glorot2010}
X.~Glorot and Y.~Bengio, ``Understanding the difficulty of training deep
  feedforward neural networks,'' in \emph{International conference on
  artificial intelligence and statistics}, 2010.

\bibitem{Jean2015}
S.~Jean, K.~Cho, R.~Memisevic, and Y.~Bengio, ``On using very large target
  vocabulary for neural machine translation,'' in \emph{arXiv preprint
  arXiv:1412.2007}, 2015.

\bibitem{Banchs2012}
R.~E. Banchs, ``Movie-dic: A movie dialogue corpus for research and
  development,'' in \emph{Proceedings of the 50th Annual Meeting of the
  Association for Computational Linguistics}, 2012, pp. 203–--207.

\bibitem{Papineni2002}
K.~Papineni, S.~Roukos, T.~Ward, and W.~Zhu, ``Bleu: A method for automatic
  evalution of machine translation,'' in \emph{Proceedings of the 40th Annual
  Meeting of the Association for Computational Linguistics}, 2002, pp.
  311–--318.

\bibitem{Lin2014}
C.~Y. Lin, ``Rouge: a package for automatic evaluation of summaries,'' in
  \emph{Proceedings of the Workshop on Text Summarization Branches Out}, 2014.

\bibitem{Li2016}
J.~Li, M.~Galley, C.~Brockett, J.~Gao, and B.~Dolan, ``A diversity-promoting
  objective function for neural conversation models,'' in \emph{Proceedings of
  NAACL-HLT}, 2016.

\end{thebibliography}

\end{document}